\documentclass{llncs}

\usepackage{microtype}
\usepackage{graphicx}
\usepackage{subfigure}
\usepackage{booktabs} 

\usepackage{hyperref}

\usepackage[font=small,labelfont=bf]{caption}
\usepackage{amssymb}
\usepackage{algorithm,algorithmic}
\usepackage{amsmath,bm}
\usepackage{epstopdf, color}
\usepackage{mathrsfs} 
\usepackage{float}

\usepackage[english]{babel}
\usepackage{blindtext}
\usepackage{multicol}

\def\BibTeX{{\rm B\kern-.05em{\sc i\kern-.025em b}\kern-.08em
    T\kern-.1667em\lower.7ex\hbox{E}\kern-.125emX}}

\begin{document}

\title{
A Framework for Following Temporal Logic Instructions with Unknown Causal Dependencies
}

\author{Duo Xu \and
Faramarz Fekri
\thanks{Accepted at IJCNN 2022 (Oral)}
}
\institute{Georgia Institute of Technology\\
 Atlanta, GA, 30332, USA}
\maketitle              


\begin{abstract}
Teaching a deep reinforcement learning (RL) agent to follow instructions in multi-task environments is a challenging problem. We consider that user defines every task by a linear temporal logic (LTL) formula. However, some causal dependencies in complex environments may be unknown to the user in advance. Hence, when human user is specifying instructions, the robot cannot solve the tasks by simply following the given instructions.
In this work, we propose a hierarchical reinforcement learning (HRL) framework in which a symbolic transition model is learned to efficiently produce high-level plans that can guide the agent efficiently solve different tasks. Specifically, the symbolic transition model is learned by inductive logic programming (ILP) to capture logic rules of state transitions. By planning over the product of the symbolic transition model and the automaton derived from the LTL formula, the agent can resolve causal dependencies and break a causally complex problem down into a sequence of simpler low-level sub-tasks.
We evaluate the proposed framework on three environments in both discrete and continuous domains, showing advantages over previous representative methods. 
\end{abstract}


\section{Introduction}
A long-standing motivation of artificial intelligence is to build agents that can understand and follow human instructions \cite{mccarthy1960programs}. Recent advances in deep reinforcement learning (RL) and language modeling have made it possible to learn a policy which produces the next action conditioned on the current observation and a natural language instruction \cite{luketina2019survey}. 
However, these approaches require manually building a large training set comprised of natural language instructions.
Recently, people have focused on using {\it formal languages} (instead of natural language) to instruct RL agents, e.g., policy sketches \cite{andreas2017modular}, reward machines \cite{icarte2022reward}, and temporal logic \cite{leon2020systematic}. These languages offer several desirable properties for RL, including clear semantics, and compact compositional syntax that enables RL practitioners to (automatically) generate massive training data to teach RL agents to follow instructions. Among popular formal languages, the linear temporal logics (LTL) is a widely-used powerful specification language for complex tasks, which allows Boolean and temporal constraints and instructions with multiple subgoals, accommodating rich specifications for many applications such as mobile robotics \cite{wongpiromsarn2010receding,maly2013iterative,vasile2013sampling}. In this work, we consider solving tasks specified by LTL instructions. 

However, some causal dependencies in complex environments may be unknown to the user in advance. Hence, when human user is specifying instructions, the robot cannot solve the tasks by simply following the given instructions.
It is well known that regular RL architectures fail in situations involving non-trivial causal dependencies that require the reasoning over an extended time horizon \cite{mnih2015human}. Therefore, solving LTL tasks can be constrained by causal dependencies unknown to the human user. This problem is not considered by the previous works on LTL task solving \cite{he2015towards,kress2018synthesis,li2021reactive,araki2021logical,vaezipoor2021ltl2action}. 


In this work, in order to tackle the problem above, we propose to first learn causal dependencies via learning symbolic operators \cite{he2015towards,silver2021learning}. Then we use a hierarchical framework to solve LTL tasks, where symbolic planning and RL are used in the high level and low level, respectively. In the high level of the proposed framework, symbolic planning is conducted in a symbolic MDP \cite{maly2013iterative,vasile2013sampling,li2021reactive} whose state is a discrete abstraction of the environment state. The symbolic operators, i.e., symbolic actions, are described in the symbolic planning domain description language (PDDL) \cite{fox2003pddl2}. In contrast to prior papers which used propositions to formulate symbolic states, here we propose to use predicates to describe relational information of objects in the environment. We first use ILP-based method to learn preconditions of symbolic operators. Then with lifted effect sets, we formulate the preconditions and effects of operators as a symbolic transition model. Combing the automaton of the given LTL formula with the learned transition model, we get a product MDP over which the symbolic planning can be conducted to produce a high-level plan of subtasks. In the low level, based on goal-conditional RL method, the policies of controllers can be trained to solve subtasks, produced by the high-level plan, one by one. 

In experiments, we conduct empirical evaluations of the proposed framework in three domains, including room, 2D reacher and 3D block stacking domains. The room domain has discrete action and state spaces, while the other two domains have continuous action and state spaces. For both training and generalization, the proposed framework is compared with representative methods from prior works.

\section{Related Work}
Recently, there has been a surge of RL papers which looked into using LTL (or similar formal languages) for reward function specification, decomposition, or shaping \cite{littman2017environment,icarte2018using,li2018policy,camacho2019ltl,yuan2019modular,jothimurugan2019composable,xu2019transfer,hasanbeig2020deep,de2020restraining,de2020temporal,jiang2020temporal}. However, all of these works formulate the LTL over propositions, without considering the relationships of objects. None of the past works consider the causal dependencies in the environment and hence, they cannot solve tasks that involve complex logical reasoning in the learning horizon.

Some past works considered investigating planning and RL in relational domain \cite{kokel2021reprel} or with causal dependencies \cite{eppe2019semantics}. However, they directly assumed that causal dependencies and objective relationships are known to the agent as prior knowledge. In our framework, instead of using these information directly, we propose to use ILP-based method to learn them. 

In the planning literature, learning symbolic transition rules have been investigated \cite{pasula2007learning,arora2018review,silver2021learning}. However, these works did not study the case in which the agent follows an instruction provided by LTL. When solving tasks that involve causal dependencies, it is necessary to learn symbolic transition rules for reasoning over a long time horizon.


\section{Background and Problem Formulation}
In this section, we are going to introduce some preliminaries and formulate the problem. The introduction of the LTL formula for task specification is presented in Appendix \ref{sec:ltl}.

\subsection{Environment MDP}
The low level part of the proposed framework is working on the environment MDP. Specifically, this environment contains a set of objects $\mathcal{O}=\{o_1,\ldots,o_n\}$ and the state $s\in\mathcal{S}$ consists of attributes of all the objects, including position, velocity, and so on. 

{\bf Controller.} We assume that the robot is equipped with a set of {\it controller} $\Pi=\{\pi^1,\ldots,\pi^K\}$ representing specific skills in the environment MDP, which are learned to solve subtasks in the low level of the proposed framework. For example, in block stacking tasks with $N$ blocks, and there are two controllers $\pi^1$ and $\pi^2$ denoting the skills of picking up object $o$ and placing object $o$ to the goal position $g$, respectively.


\subsection{Symbolic MDP}
\label{sec:rmdp}
In prior related papers \cite{icarte2018using,illanes2020symbolic,araki2021logical}, the high-level part of the environment is represented by a symbolic MDP, where the state and action are described by propositions which ignore the relations of objects. In this work, the high level is defined as a the relational domain specified by objects in $\mathcal{O}$ and a set of predicates representing relationships of objects and events in the environment. A predicate with only variables as arguments is called {\it lifted}; a predicate with objects as arguments is called as a {\it grounded predicate} or {\it atom}. Each predicate is a classifier over the environment state $s$. All the grounded predicates are assumed to be in the set $\mathcal{Q}$ whose values (True or False) are determined by a deterministic labeling function $L$. Given the environment state $s$, we can compute the atoms that hold true in the state $s$ by $L$. And the output of function $L$ formulates the {\it symbolic state}, i.e., $\tilde{s}:=L(s)\in2^{\mathcal{Q}}$. 

{\bf Symbolic MDP} Formally, a symbolic MDP is defined as $\tilde{\mathcal{M}}:=\langle \tilde{\mathcal{S}}, \tilde{\mathcal{A}}, \tilde{s}_0, \tilde{T}, \tilde{R}, \gamma\rangle$. The {\it symbolic states} in $\tilde{\mathcal{S}}$ are subsets of atoms (grounded predicates) in $\mathcal{Q}$. The {\it symbolic action} in $\tilde{\mathcal{A}}$ is a PDDL-style operator predicate \cite{fox2003pddl2} grounded by objects in $\mathcal{O}$. In addition, $\tilde{s}_0$ is the initial symbolic state, $\tilde{T}$ is a probability {\it transition function} $\tilde{\mathcal{S}}\times \tilde{\mathcal{A}}\times \tilde{\mathcal{S}}\to[0, 1]$ which is {\it unknown a priori}, $\tilde{R}$ is the reward function, and $\gamma\in[0,1)$ is the discount factor. We also distinguish the set of subgoal atoms as $\mathcal{Q}_G\subset\mathcal{Q}$ which are used to define the atoms of the LTL formula. 

{\bf Operators.} A PDDL operator $\text{op}\in\mathcal{OP}$ is defined as a lifted predicate with objects as arguments, playing the role of symbolic action in the high level, e.g., in block stacking, the operator Place(X,Y) refers to moving the object X onto the top of Y. A grounded operator means the operator with arguments replaced by objects in $\mathcal{O}$, defining a symbolic action in $\tilde{\mathcal{M}}$. Formally, an operator $\text{op}$ is defined by a tuple $\langle p(\text{op}), \text{pre}(\text{op}), \text{eff}(\text{op})\rangle$ \cite{silver2021learning}, where $p$ is the corresponding controller, and $\text{pre}, \text{eff}$ are logic rules for preconditions and effects of the operator $\text{op}$, respectively. Preconditions are lifted predicates that describe what must hold for the applicability (causal dependencies) of the operator. Effects are lifted predicates that describe how the symbolic state in $\tilde{\mathcal{S}}$ changes as a result of applying this operator successfully, which consists of positive and negative effects. The termination of the operator is determined by its controller policy in the low level, i.e., when $p(\text{op})$ successfully reaches its goal, the operator op terminates. 

\subsection{Inductive Logic Programming}
\label{sec:pre_ilp}
We use Inductive logic programming (ILP) to derive a definition (set of clauses) of some lifted predicates, given some positive and negative examples \cite{koller2007introduction,evans2018learning}. Conducting ILP with differentiable architectures has been investigated in many previous work \cite{evans2018learning,rocktaschel2017end,dong2019neural,payani2019learning}. In this work we use an ILP-based method to learn the symbolic transition rules.

\subsection{Problem Formulation}
In this work, the problem is to train a robot to follow LTL instructions in a relational domain that involve complex causal dependencies among symbolic operators (actions in symbolic MDP $\tilde{M}$). Different from prior papers, we assume that causal dependencies of operators, i.e., symbolic transition rules, are unknown a priori and not contained in the LTL formula. In order to overcome the long planning horizon and sparsity of rewards, we propose to decompose the original problem into high-level and low-level parts which are solved by symbolic planning and goal-conditioned RL separately. 

\section{Methodology}
\begin{figure}
    \vspace*{-10pt}
    \centering
    \fontsize{8pt}{10pt}\selectfont
    \def\svgwidth{0.6\columnwidth}
    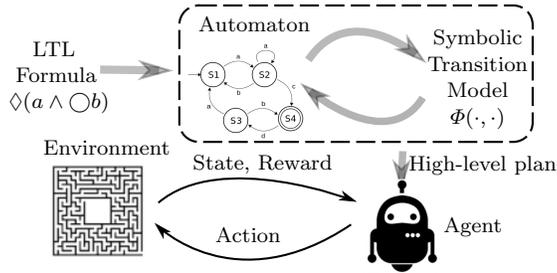
    \caption{Diagram of the proposed framework. The Automaton defines the logic FSA $\mathcal{A}_{\phi}$. The symbolic transition model $\Phi$ approximates the transition rules of symbolic MDP $\tilde{\mathcal{M}}$, i.e., the precondition and effects of operators. The dashed rectangle represents the product MDP $\mathcal{P}$.}
    \label{fig:hrl}
    \vspace*{-10pt}
\end{figure}

In order to solve the problems defined above, we propose the framework described in Figure \ref{fig:hrl}. 
In the high level, the environment is abstracted into a symbolic MDP $\tilde{\mathcal{M}}$ which encodes the complex causal dependencies in the environment, whereas in the low level, the controllers $\Pi$ of the environment are realized by goal-conditioned policies. 

The proposed method has two stages. First, the symbolic transition model $\Phi$ is learned by an ILP-based method to model causal dependencies of symbolic operators, while the policy of every low-level controller in $\Pi$ is trained by the goal-conditioned RL method. In the second stage, when learning to solve a task, the LTL formula $\phi$ for task specification is first converted into a finite state automaton (FSA) $\mathcal{A}_{\phi}$. Then over the product MDP of $\Phi$ and $\mathcal{A}_{\phi}$, a high-level plan $h$, which satisfies both LTL specifications and causal dependencies, can be found by a symbolic planner based on logic value iterations. Hence, the agent can use low-level policies of controllers to finish grounded operators in $h$ one by one, which can solve the whole LTL task successfully. The introduction of LTL and FSA is presented in Appendix \ref{sec:ltl}.


\begin{figure}
    \vspace*{-20pt}
    \centering
    \fontsize{8pt}{10pt}\selectfont
    \def\svgwidth{0.5\columnwidth}
\begingroup%
  \makeatletter%
  \providecommand\color[2][]{%
    \errmessage{(Inkscape) Color is used for the text in Inkscape, but the package 'color.sty' is not loaded}%
    \renewcommand\color[2][]{}%
  }%
  \providecommand\transparent[1]{%
    \errmessage{(Inkscape) Transparency is used (non-zero) for the text in Inkscape, but the package 'transparent.sty' is not loaded}%
    \renewcommand\transparent[1]{}%
  }%
  \providecommand\rotatebox[2]{#2}%
  \newcommand*\fsize{\dimexpr\f@size pt\relax}%
  \newcommand*\lineheight[1]{\fontsize{\fsize}{#1\fsize}\selectfont}%
  \ifx\svgwidth\undefined%
    \setlength{\unitlength}{226.80958484bp}%
    \ifx\svgscale\undefined%
      \relax%
    \else%
      \setlength{\unitlength}{\unitlength * \real{\svgscale}}%
    \fi%
  \else%
    \setlength{\unitlength}{\svgwidth}%
  \fi%
  \global\let\svgwidth\undefined%
  \global\let\svgscale\undefined%
  \makeatother%
  \begin{picture}(1,0.13728229)%
    \lineheight{1}%
    \setlength\tabcolsep{0pt}%
    \put(0,0){\includegraphics[width=\unitlength,page=1]{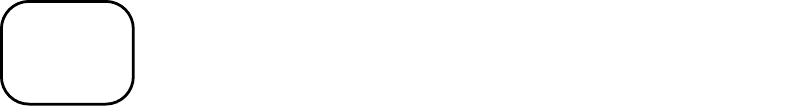}}%
    \put(0.05773067,0.05573124){\color[rgb]{0,0,0}\makebox(0,0)[lt]{\lineheight{1.25}\smash{\begin{tabular}[t]{l}$D$\end{tabular}}}}%
    \put(0,0){\includegraphics[width=\unitlength,page=2]{dataset_op.pdf}}%
    \put(0.32748192,0.05538016){\color[rgb]{0,0,0}\makebox(0,0)[lt]{\lineheight{1.25}\smash{\begin{tabular}[t]{l}$D_{op}$\end{tabular}}}}%
    \put(0,0){\includegraphics[width=\unitlength,page=3]{dataset_op.pdf}}%
    \put(0.57900063,0.05244276){\color[rgb]{0,0,0}\makebox(0,0)[lt]{\lineheight{1.25}\smash{\begin{tabular}[t]{l}$D_{op,m}$\end{tabular}}}}%
    \put(0,0){\includegraphics[width=\unitlength,page=4]{dataset_op.pdf}}%
  \end{picture}%
\endgroup%

    \caption{The process of learning symbolic transition model $\Phi$: 1) collect symbolic transition data; 2) partition the dataset by operator predicate; 3) learn effects by clustering the dataset further; 4) learn precondition by logic neural network.}
    \label{fig:ilp}
    \vspace*{-10pt}
\end{figure}

\subsection{Learning Symbolic Transition Model}
\label{sec:symb_op}

The symbolic transition model $\Phi$ is to predict the next symbolic state $\tilde{s}_{t+1}$ given the current symbolic state $\tilde{s}_t$ and action (grounded operator) $\tilde{a}_t$. This model $\Phi$ consists of preconditions and effects of operators. For each operator op$\in\mathcal{OP}$, if $\tilde{s}_t$ satisfies pre(op), the effects eff(op) would be applied onto $\tilde{s}_t$ for predicting $\tilde{s}_{t+1}$, with atoms in $\tilde{s}_t$ added or removed. 

In our framework, based on partitioned dataset of transitions in symbolic MDP $\tilde{\mathcal{M}}$, we use clustering to learn {\it effects} in the form of lifted predicates, and use ILP-based method to learn {\it preconditions}, which can be combined to build the symbolic transition model $\Phi$. The general process of learning $\Phi$ is summarized in Figure \ref{fig:ilp}.

{\bf Dataset Partitioning.} We first collect a dataset $D=\{(\tilde{s}_i, \tilde{a}_i,\tilde{s}_{i+1})\}$ of symbolic transition tuples in $\tilde{\mathcal{M}}$, which is used to learn effects and preconditions of operators (symbolic actions). The method of data collection in $\tilde{\mathcal{M}}$ may vary across different environments and the details are introduced in Section \ref{sec:exp}. Then the dataset $D$ is partitioned according to the operator predicates of action $\tilde{a}_i$. E.g., in block stacking tasks, $D$ can be partitioned into $D_{\text{pick}}$ and $D_{\text{place}}$, corresponding to operator predicates Pick($\cdot$) and Place($\cdot,\cdot$) respectively. 

Then, for every operator $op\in\mathcal{OP}$, we further cluster the transition data $D_{op}$ according to lifted effects, which can learn the {\it effects} of applying operators in different symbolic states. 
Specifically, for any transition tuple $(\tilde{s}_i, \tilde{a}_i, \tilde{s}_{i+1})$ in $D_{op}$, the {\it grounded effects} of applying operator $op$ include positive and negative effects which are computed as $\tilde{s}_{i+1}-\tilde{s}_i$ and $\tilde{s}_i-\tilde{s}_{i+1}$, representing added and removed atoms in this state transition, respectively. We then cluster pairs of transitions together if their effects can be {\it unified}, that is, if there exists a bijective mapping between the objects in the two transitions such that the effects are equivalent up to this mapping \cite{icarte2018using,illanes2020symbolic,kokel2021reprel}. Each of the resulting clusters is indexed with the {\it lifted effect} set, where the objects in the same effect set from any arbitrary one of the constituent transitions are replaced with variables. For operator op$\in\mathcal{OP}$, we denote the clustered dataset of transitions for $m$-th lifted effect set as $D_{\text{op}, m}$. The lifted effects obtained by clustering $D_{op}$ tell us the effects of operator $op$ in different symbolic states. 

Note that if the execution of operator $\tilde{a}_i$ is not successful (operator's precondition is not satisfied), we will have $\tilde{s}_i=\tilde{s}_{i+1}$, where the lifted effect set is empty. The clustered dataset for this none effect is specifically denoted as $D_{\text{op}, 0}$. 


For every $D_{\text{op}, m} (m>0)$, we apply an ILP-based method to learn the precondition for the $m$-th effect set of the operator op. The details of the method of learning $\Phi$ is presented in Algorithm \ref{alg:ilp} in Appendix, which is summarized in Figure \ref{fig:ilp}.

{\bf ILP-based Method to Learn Preconditions. } We adopt the logic neural network \cite{wang2021scalable,payani2019learning} to learn logic rules for preconditions of symbolic operators, where rules can be represented conjunctive normal forms and made differentiable by using logic activation function. In logic neural networks, the input layer consists of Boolean-valued atoms (grounded predicates) of the symbolic state. Then, we use one logic layer consisting of one conjunction layer and one disjunction layer, and every node in the logic layer represents one rule. The logic layer and input layer are connected by a trainable weight matrix $W$ where predicates are grounded by the different objects sharing same weights. The output layer, which is a linear layer, only has one binary output which denotes the applicability of the operator in the input symbolic state. 

In the training, we use the continuous version of the logic layer for optimization \cite{payani2019learning,wang2021scalable}. The conjunction and disjunction functions are defined as $F_c(x,\omega):=1-\omega(1-x)$ and $F_d(x,\omega):=x\cdot\omega$, respectively. And we use the logic activation function $\mathbb{P}(v)=\frac{-1}{-1+\log(v)}$ from \cite{wang2021scalable}. Specifically, given the vector of atom values $\bm{x}$, the rule at the $i$-node of the conjunction and disjunction layer can be expressed as
\begin{eqnarray}
    \text{Conj}_i(\bm{x}, W_i)&:=&\mathbb{P}(\prod_{j=1}^n(F_c(x_j, W_{i,j})+\epsilon)), \nonumber \\ \text{Disj}_i(\bm{x}, W_i)&:=&1-\mathbb{P}(\prod_{j=1}^n(1-F_d(x_j, W_{i,j})+\epsilon)) \nonumber \label{logic_layer}
\end{eqnarray}
where $W_i$ is the trainable weight vector for the $i$-th node in the logic layer, and $\epsilon$ is a small constant, e.g., $10^{-10}$, for numerical stability.



\subsection{Learning Controller Policies}
\label{sec:op_to_act}
In order to map the grounded operators in the high-level plan into action sequences of the low level, we adopt the goal-conditioned reinforcement learning (GCRL) approach \cite{florensa2018automatic,nair2018visual}. For every controller $\pi^k\in\Pi$, we train a goal-conditioned policy which aims at reaching an independent goal with high probability. For instance, in block stacking tasks, the controller $\pi^1$ refers to moving the robotic arm to pick up a block whose position determines the goal for the policy. 
The details of learning GCRL policies are presented in Algorithm \ref{alg:sac-her} in Appendix.

\subsection{Symbolic Planning}
\label{sec:plan}
{\bf Product MDP.} The high-level part of the proposed framework is to find a plan consisting of a sequence of grounded operators which can then be executed by low-level controller policies to finish the LTL task. In order to find plans that satisfy both the LTL specification $\phi$ and causal dependencies (preconditions) of operators, we construct a {\it product MDP} $\mathcal{P}$ \cite{he2015towards,li2021reactive} of the symbolic MDP $\tilde{\mathcal{M}}$ and the FSA $\mathcal{A}_{\phi}$, i.e., $\mathcal{P}=\tilde{\mathcal{M}}\times\mathcal{A}_{\phi}$, as shown in Figure \ref{fig:hrl}. We use the learned model $\Phi$ as the transition function of $\tilde{\mathcal{M}}$. The definition of FSA $\mathcal{A}_{\phi}$ is introduced in Appendix \ref{sec:ltl}. Specifically, we define $\mathcal{P}=\langle\mathcal{Z}_p, z_{p,0}, \Sigma, T_p, \mathcal{F}_p\rangle$, where $\mathcal{Z}_p=\tilde{\mathcal{S}}\times\mathcal{Z}_a$ whose element is a tuple of $(\tilde{s}, z_a)\in\mathcal{Z}_p$, $z_{p,0}=(\tilde{s}_0, z_{a,0})$, $\Sigma=2^{\mathcal{Q}}$, $T_p:\mathcal{Z}_p\times\mathcal{Z}_p\to[0,1]$, and $\mathcal{F}_p\subset\mathcal{Z}_p$. Note that the product transition function satisfies $(\tilde{s}', z_a')\sim T_p(\cdot|\tilde{s}, z_a)$ iff there exists $\tilde{a}\in\tilde{\mathcal{A}}$ and $T_a$ such that $\tilde{s}'\sim\Phi(\cdot|\tilde{s}, \tilde{a})$ and $z'\sim T_a(\cdot|z_a, \tilde{s}')$, where $T_a$ is the state transition function of FSA $\mathcal{A}_{\phi}$.

In the high level, we use $\mathcal{P}$ as a search graph for finding a plan, which is to find a valid trajectory starting from an initial state $(\tilde{s}_0, z_{a,0})$ to one of the accepting states $(\tilde{s}_K, z_{a,F})$ where $z_{a,F}\in\mathcal{F}_a$ in FSA $\mathcal{A}_{\phi}$, i.e., $\xi:=\{(\tilde{s}_i, z_{a,i}, \tilde{a}_i)\}_{i=0}^K$, where $\tilde{a}_i$ satisfies $\tilde{s}_{i+1}\sim\Phi(\cdot|\tilde{s}_i, \tilde{a}_i)$ and $z_{a,i+1}\sim T_a(\cdot|z_{a,i},\tilde{s}_{i+1})$. Then, a high-level plan $h:=\{\tilde{a}_i\}_{i=0}^{K-1}$ consisting of a sequence of grounded operators can be directly extracted from the valid trajectory $\xi$.

Specifically, in finding $\xi$, we construct a graph $G=\langle V, E, \omega\rangle$ and use the symbolic planning algorithm (logic value iteration \cite{araki2019learning}) to find a valid trajectory. In particular, every state $z_p\in\mathcal{Z}_p$ corresponds to a node $v\in V$. And there is an edge $e\in E$ connecting the pair $z_p=(\tilde{s},z_{a})$ and $z_p'=(\tilde{s}',z_{a}')$ where $z_p'\sim T_p(\cdot|z_p), \tilde{s}\neq\tilde{s}'$ and there exists $\tilde{a}$ satisfying $\tilde{s}'\sim\Phi(\cdot|\tilde{s},\tilde{a})$. The cost of the edge $\omega(z_p, z_p')$ is read from the critics of the low-level controller for the operator $\tilde{a}$.  

\subsection{Integrating Symbolic Planning and Reinforcement Learning}
\label{sec:int}
With learned symbolic transition model $\Phi$, the proposed hierarchical framework integrates symbolic planning in the high level with goal-conditioned RL in the low level, as shown in Figure \ref{fig:hrl}. We can generate valid high-level plans over this learned transition model $\Phi$ and FSA $\mathcal{A}_{\phi}$, without applying actions/operators in the practical environment, so that we can learn and follow causal dependencies in a sample-efficient manner.  

For every grounded operator $\tilde{a}$ in the plan $h$, we select the corresponding controller policy in the low level whose goal is grounded by the positions of objects. For instance, in block stacking tasks, for Place($o_1, o_2$), the controller policy for placing $\pi^2$ is selected and the goal is set to be the position of object $o_2$. The details of the proposed framework are in Algorithm \ref{alg1}. 

\begin{algorithm}[ht]
\caption{The Proposed Framework for Hierarchical RL with LTL Objectives}
\label{alg1}
\begin{algorithmic}[1]
\REQUIRE Environment MDP $\mathcal{M}=\langle\mathcal{S}, \bm{s}_0, \mathcal{A}, \mathcal{T}\rangle$; \\
symbolic MDP $\tilde{\mathcal{M}}=\langle\tilde{\mathcal{S}}, \tilde{\mathcal{A}}, \tilde{s}_0, \tilde{T},\tilde{R},\gamma\rangle$; \\
The set of grounded predicates (atoms) $\mathcal{Q}$, the set of subgoal atoms $\mathcal{Q}_G\subset\mathcal{Q}$, the set of objects $\mathcal{Q}$, the set of operator predicates $\mathcal{OP}$, and the labeling function $L:\mathcal{S}\to\tilde{\mathcal{S}}$; \\
The LTL formula $\phi$ for task specification given by the user; \\
The FSA of the given LTL formula $\mathcal{A}_{\phi}=\langle\mathcal{Z}_a, z_{a,0}, \Sigma, T_a, \mathcal{F}_a\rangle$; \\
\STATE Apply Algorithm \ref{alg:sac-her} in Appendix to learn low-level controller policies in $\Pi$;
\STATE Apply Algorithm \ref{alg:ilp} in Appendix to learn the symbolic transition model $\Phi$;
\STATE {\it //Find a high-level plan $h$ over the product MDP $\mathcal{P}$ by Logic Value Iteration:}
\STATE Initialize $Q:\mathcal{Z}_a\times\tilde{\mathcal{S}}\times\tilde{\mathcal{A}}\to\mathbb{R}, V:\mathcal{Z}_a\times\tilde{\mathcal{S}}\to\mathbb{R}$ to $0$;
\FOR{$k=1,\ldots,K$}
\FOR{$(z,\tilde{s})\in\mathcal{Z}_a\times\tilde{\mathcal{S}}$}
\FOR{$\tilde{a}\in\tilde{\mathcal{A}}$}
\STATE $Q_k(z,\tilde{s},\tilde{a})\leftarrow \tilde{R}(\tilde{s})+\sum_{(z', \tilde{s}')\in\mathcal{Z}_a\times\tilde{\mathcal{S}}}\Phi(\tilde{s}'|\tilde{s},\tilde{a})T_a(z'|z,\tilde{s}')V_{k-1}(z', \tilde{s}')$
\ENDFOR
\STATE $V_k(z,\tilde{s})\leftarrow\max_{\tilde{a}\in\tilde{\mathcal{A}}}Q_k(z,\tilde{s},\tilde{a})$
\ENDFOR
\ENDFOR
\STATE Initialize $h\leftarrow\{\}, \tilde{s}\leftarrow\tilde{s}_0, z\leftarrow z_{a,0}$;
\WHILE{$z\not\in\mathcal{F}_a$}
\STATE $\tilde{a}\leftarrow\arg\max_{a'\in\tilde{\mathcal{A}}}Q_K(z,\tilde{s},a')$;
\STATE Append $\tilde{a}$ to $h$;
\STATE Update $\tilde{s}\sim \Phi(\cdot|\tilde{s},\tilde{a})$ and then $z\sim T_a(\cdot|z,\tilde{s})$;
\ENDWHILE
\STATE {\bf Return} Controller policies $\Pi$ and high-level plan $h$
\end{algorithmic}
\end{algorithm}

\section{Experiments}
\label{sec:exp}

The method of learning symbolic transition model $\Phi$ is evaluated in three environments, including room, reacher and block stacking domains. For LTL task solving, we conducted experiments to evaluate the proposed framework on the {\it training} of the given LTL instruction and the {\it generalization} to other ones. Since the LTL task has multiple subgoals, the evaluation of training and generalization should be separated. For training, the LTL formulae are chosen to be randomly generated, and the metric of evaluation is success rate. For generalization, given a random LTL task, based on the trained controller policies and value functions, the number of re-training steps for searching a high-level plan is recorded and compared with baselines. The metric in generalization is the success rate of testing tasks. 
\\
{\bf Environments.} The performance of the proposed framework is evaluated on three environments (domains).
\begin{itemize}
    \item Room domain: the robot is tasked to visit several rooms in a specific temporal order. In this environment, the robot has to pick up keys and open the doors and the door can only be opened by keys in the same color, which imposes causal dependencies of visiting different rooms. The task in this domain is to visit multiple rooms in an order satisfying the LTL formula.
    \item Reacher domain: it is a two-link arm that has continuous state and action spaces. There are multiple objects denoted as colored balls, such as green $g$, red $r$, blue $b$, yellow $y$, white $w$. In order to introduce causal dependencies, we impose pre-conditions of visiting some balls. For example, red ball can only be visited after yellow ball is visited, and red and blue balls have to be visited before the robot goes to the green one. The task in this domain is to visit multiple colored balls in a valid order. 
    \item Block stacking domain: it is a continuous adaptation of the classical blocks world problem simulated by PyBullet \cite{coumans2016pybullet}. A robot uses its gripper to interact with blocks on a tabletop and must assemble them into various towers. The robot has two operators, Pick and Place, which are to pick up certain block and place it on top of another block, respectively. There are many causal dependencies here, e.g., the gripper has to be empty before picking up a block, and the top of target block should be empty before the robot places something there. The task in this domain is to realize different On$(o_1, o_2)$ in an order satisfying the given LTL formula.
\end{itemize}

{\bf Task.} We evaluate the proposed framework on three tasks. Every task is randomly generated, and the robot does not know any information about the task before learning starts. The first task is "sequential" task and is written in the form of $\phi_{\text{seq}}=\Diamond(a\wedge\Diamond(b\wedge\Diamond(c\wedge\Diamond d)))$, where the atom has to be realized in the order of $a,b,c$ and $d$, and atoms are selected randomly. The length of sequential formula (number of atoms) are randomly selected between 2 and 5. 
The second task is "OR" task that concatenates terms by disjunctive operator $\vee$ where every term is a short sequential task, e.g., $\phi_{\text{or}}=\Diamond(a\wedge\Diamond b)\vee\Diamond(c\wedge\Diamond d)$. Specifically, the number of terms is ranging between 2 and 4, and the length of every term is from 1 to 3. 
The third task is called "recursive" task \cite{vaezipoor2021ltl2action}, which can be formulated as $\phi_{\text{rec}}=\phi_{\text{rec}}\wedge\phi'|\phi'$, $\phi'=\Diamond(p'\wedge\phi')|\Diamond p'$, and $p'=a|a\vee b$, where $a$ and $b$ are atoms for subgoals. The notation $|$ is for alternative, and in task formula generation, two sub-formulae around $|$ are uniformly selected. The depth of recursion is randomly selected between 1 and 3. 
\\
{\bf Baselines.} In order to test the effects of the learned symbolic transition model, we evaluate two baselines against the proposed framework. The baselines considered here still use hierarchical RL to solve given tasks, but they use different methods from the proposed framework in the high-level part, where the symbolic transition rules are not learned or utilized. 
\begin{itemize}
    \item Baseline-1: The first baseline is to use {\it Q-learning} to find a high-level plan, which does not need to utilize the transition rules in either symbolic model $\Phi$ or FSA $\mathcal{A}_{\phi}$. 
    \item Baseline-2: The second baseline uses {\it Reward Machines} \cite{icarte2018using} to find the high-level plan, where the Q-value functions are trained for reaching every state in the FSA $\mathcal{A}_{\phi}$. In this approach, the original task is decomposed into sub-tasks of reaching different states of $\mathcal{A}_{\phi}$ and solved independently by Q-learning. 
\end{itemize}
The details and implementation of baselines are presented in Appendix \ref{sec:base}.
\\
{\bf Data Collection for Learning $\Phi$.} In order to learn the symbolic transition model $\Phi$, in every domain, we use random high-level policies to collect symbolic transition data for $K$ trajectories, denoted as $\mathcal{D}$, where every grounded operator is uniformly selected and the maximum length of each trajectory is 100. Specifically, for data collection, the room domain has $4\times4$ rooms with two pairs of keys and locks where $K=50$. The reacher domain and block stacking domains have the same number of objects as those in the evaluation where $K=100$. In every domain, the low-level controller policies are pre-trained before data collection, and the agent will be reset to its previous state if any grounded operator is not successfully executed.


\subsection{Learning Symbolic Transition Model}
\subsubsection{Room}
\label{sec:room}
{\bf Setup.} In this domain, the symbolic state is defined by the following predicates: At(X), Visited(X), Connect(X,Y), Lock(X,Y,C) and hasKey(C), RoomHasKey(X,C). The predicates in the symbolic state, At(X) and Visited(X), denote the current room and previously visited rooms of the robot. The Connect(X,Y) means there is a corridor between room X and Y. Lock(X,Y,C) denotes a lock in color C between room X and Y. The predicates hasKey(C) and RoomHasKey(X,C) denote the key in color C is at the robot or room X, respectively. The operator predicate is FromTo(X,Y), which is also the predicate for representing symbolic actions. Here X and Y are variables of room indices. Since only neighboring rooms are connected by corridors or locks, the low-level controller policy is only to visit next rooms in 4 cardinal directions, which is so simple that GCRL policy is not used here. The application of operator can be blocked by walls and locks between rooms.
\\
{\bf Learned Transition Rules.} By applying the learning method introduced in Section \ref{sec:symb_op} on the dataset $\mathcal{D}$, we can learn the transition rules for operator predicates as below:
\\
\\
\noindent\rule[0.25\baselineskip]{\columnwidth}{1pt}
\begin{small}
\begin{itemize}
\vspace*{-5pt}
    \item FromTo(X,Y): 
    \begin{itemize}
        \item {\bf precondition}: At(X), Connect(X,Y)
        \item {\bf effect}: At(Y), Visited(Y)
    \end{itemize}
    \item FromTo(X,Y):
    \begin{itemize}
        \item {\bf precondition}: {At}(X), {Connect}(X,Y), RoomHasKey(X,C)
        \item {\bf effect}: At(Y), {Visited}(Y), {hasKey}(C), $\neg${RoomHasKey}(X,C)
    \end{itemize}
    \item FromTo(X,Y): 
    \begin{itemize}
        \item {\bf precondition}: {At}(X), {Lock}(X,Y,C), {hasKey}(C)
        \item {\bf effect}: {At}(Y), {Visited}(Y), {Connect}(X,Y), $\neg${Lock}(X,Y,C)
    \end{itemize}
\vspace*{-5pt}
\end{itemize}
\end{small}
\noindent\rule[0.25\baselineskip]{\columnwidth}{1pt}

\subsubsection{Reacher}
\label{sec:reacher}
{\bf Setup.} In the reacher domain, the symbolic state consists of RedVisited(), BlueVisited(), and so on, meaning that the ball in some color has been visited or not. The operator is GoRed(), GoBlue(), and so on, denoting visiting a ball with a specific color. The low-level policies of controller is trained by GCRL, where the goal space is defined as the $(x,y)$ coordinates of the target ball. The visiting of some balls is constrained by partial orders, e.g., visiting red (blue) ball is necessary before visiting the green (yellow) ball.
\\
{\bf Learned Transition Rule.} Based on the learning method in Section \ref{sec:symb_op}, we learn the following transition rules from the collected data. Operators without constraints are omitted here\\
\noindent\rule[0.25\baselineskip]{\columnwidth}{1pt}
\begin{small}
\begin{itemize}
\vspace*{-5pt}
    \item GoToGreen():
    \begin{itemize}
        \item {\bf{precondition}}: {RedVisited}()
        \item {\bf effect}: GreenVisited()
    \end{itemize}
    \item GoToYellow():
    \begin{itemize}
        \item {\bf precondition}: BlueVisited()
        \item {\bf effect}: YellowVisited()
    \end{itemize}
\vspace*{-5pt}
\end{itemize}
\end{small}
\noindent\rule[0.25\baselineskip]{\columnwidth}{1pt}

\subsubsection{Block Stacking}
\label{sec:block}
{\bf Setup.} In the definition of symbolic state space $\tilde{\mathcal{S}}$, the predicates are On($o_1, o_2$), TopEmpty($o$), OnTable($o$), Holding($o$) and GripperEmpty(), meaning that the object $o_1$ is on top of $o_2$, object $o$ has nothing on top of it, object $o$ is on the table, the gripper is holding object $o$, and the gripper holds nothing, respectively. There are three operator predicates, Pick($o$), Place($o_1, o_2$) and PutOnTable($o$), formulating symbolic actions. The goal vector $g$ can be easily obtained from the positions of related objects.  In the low level, there are two controller policies (skills of the robot), $\pi^{\text{pick}}(a|s,g)$ and $\pi^{\text{place}}(a|s,g)$, which are to pick up and place some object to some place by the gripper, respectively. The operator Pick($o$) is accomplished by the controller $\pi^{\text{pick}}(a|s,g)$, whereas the operators Place($o_1, o_2$) and PutOnTable($o$) are accomplished the controller $\pi^{\text{place}}(a|s,g)$.
\\
\\
{\bf Learned Transition Rules.} Based on the collected symbolic transition data $D$, we can learn first-order rules describing the pre-condition and effect of operator predicates as below,
\\
\\
\noindent\rule[0.25\baselineskip]{\columnwidth}{1pt}
\begin{small}
\begin{itemize}
    \item Pick(X):
    \begin{itemize}
        \item {\bf precondition}: {GripperEmpty}(), {TopEmpty}(X)
        \item {\bf effects}: {Holding}(X), $\neg${GripperEmpty}()
    \end{itemize}
    \item Place(X,Y):
    \begin{itemize}
        \item {\bf precondition}: {Holding}(X), {TopEmpty}(Y), {On}(X,Z)
        \item {\bf effects}: {GripperEmpty}(), {On}(X,Y), {TopEmpty}(Z), $\neg${Holding}(X), $\neg${TopEmpty}(Y), $\neg${On}(X,Z)
    \end{itemize}
    \item Place(X,Y):
    \begin{itemize}
        \item {\bf precondition}: {Holding}(X), {TopEmpty}(Y), {OnTable}(X)
        \item {\bf effects}: {GripperEmpty}(), {On}(X,Y), $\neg${Holding}(X), $\neg${TopEmpty}(Y), $\neg${OnTable}(X)
    \end{itemize}
    \item PutOnTable(X):
    \begin{itemize}
        \item {\bf precondition}: {Holding}(X), {On}(X, Y)
        \item {\bf effects}: {GripperEmpty}(), {OnTable}(X), $\neg${Holding}(X), $\neg${On}(X, Y)
    \end{itemize}
\end{itemize}
\vspace*{-5pt}
\end{small}
\noindent\rule[0.25\baselineskip]{\columnwidth}{1pt}

\begin{figure}[ht]
    \vspace*{-10pt}
    \centering
    \subfigure[Room]{
    \centering
    \includegraphics[width=1.15in, height=.9in]{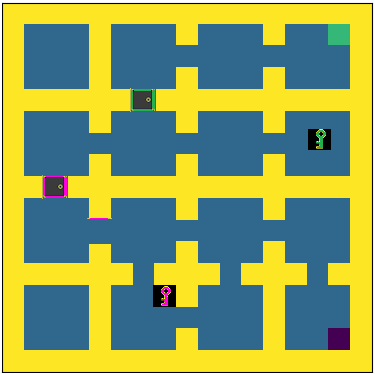}
    \label{fig:result_room_1}
    }
    \subfigure[Training]{
    \centering
    \includegraphics[width=1.35in]{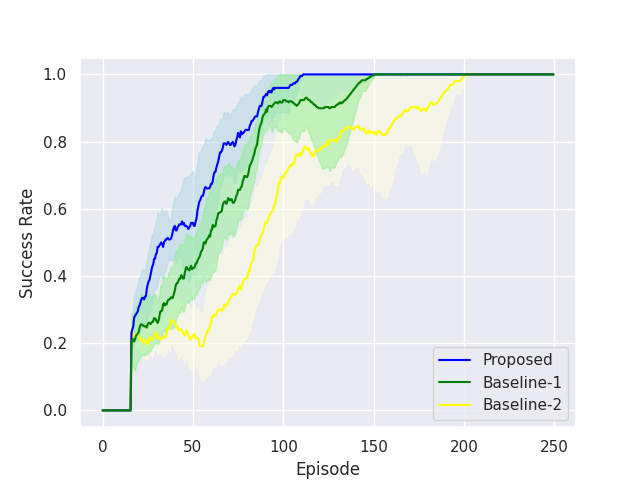}
    \label{fig:result_room_2}
    }
    \subfigure[Generalization]{
    \centering
    \includegraphics[width=1.35in]{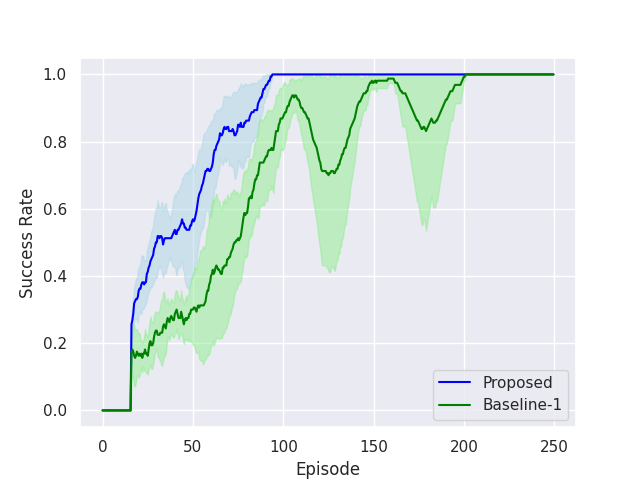}
    \label{fig:result_room_3}
    }
    \vspace*{-10pt}

    \subfigure[Reacher]{
    \centering
    \includegraphics[width=1.15in, height=.9in]{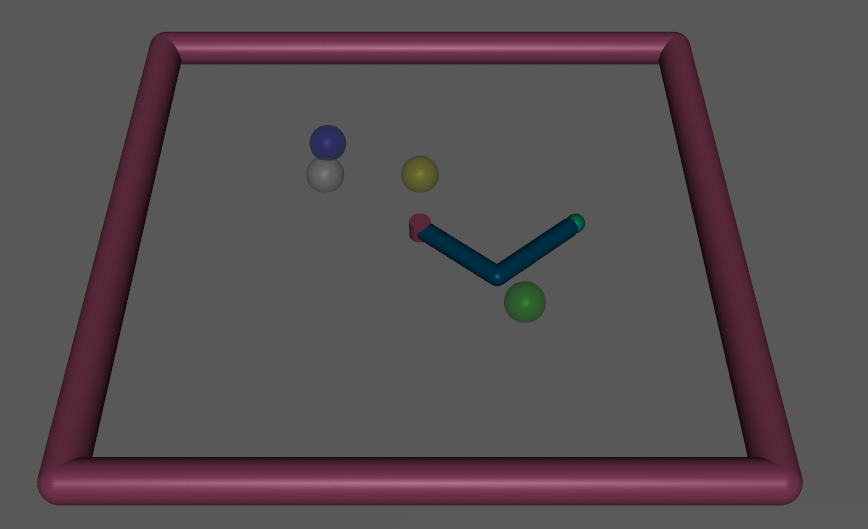}
    \label{fig:result_reacher_1}
    }
    \subfigure[Training]{
    \centering
    \includegraphics[width=1.35in]{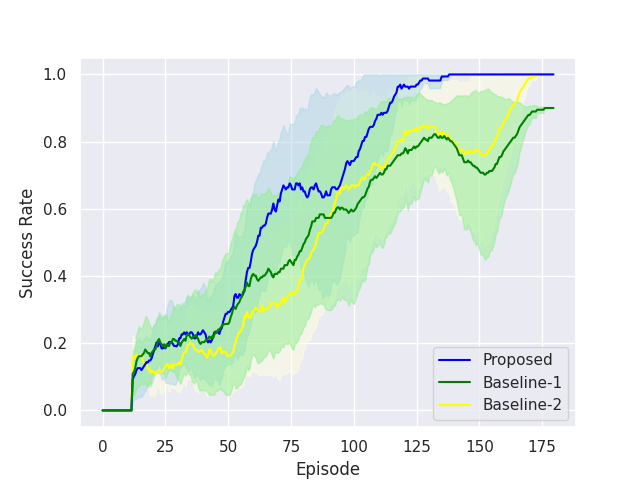}
    \label{fig:result_reacher_2}
    }
    \subfigure[Generalization]{
    \centering
    \includegraphics[width=1.35in]{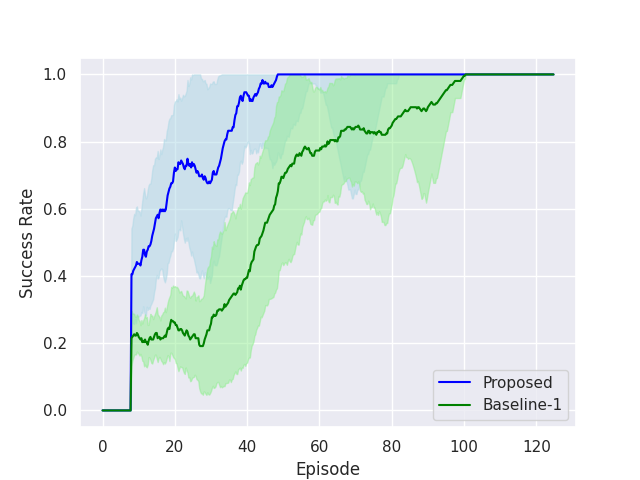}
    \label{fig:result_reacher_3}
    }
    \vspace*{-10pt}

    \subfigure[Block Stacking]{
    \centering
    \includegraphics[width=1.15in, height=.9in]{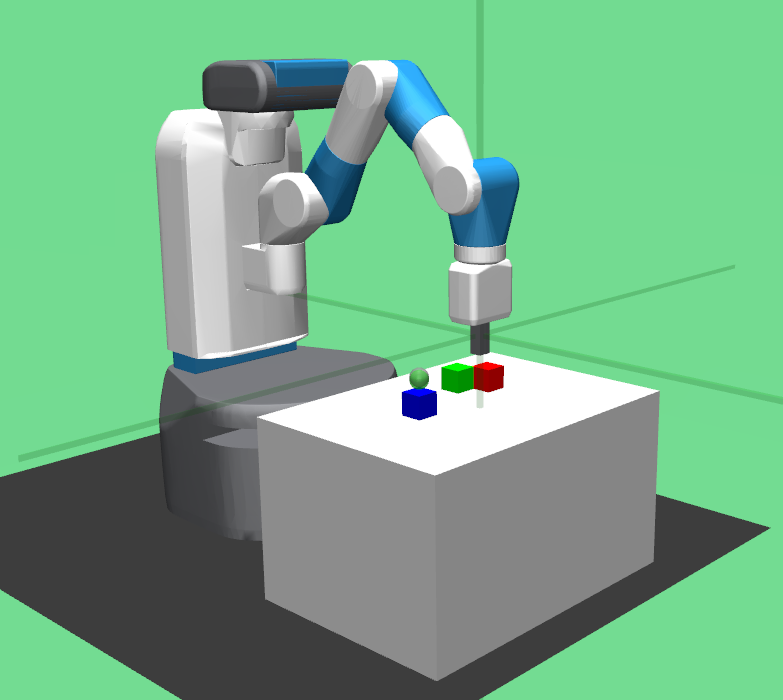}
    \label{fig:result_block_1}
    }
    \subfigure[Training]{
    \centering
    \includegraphics[width=1.35in]{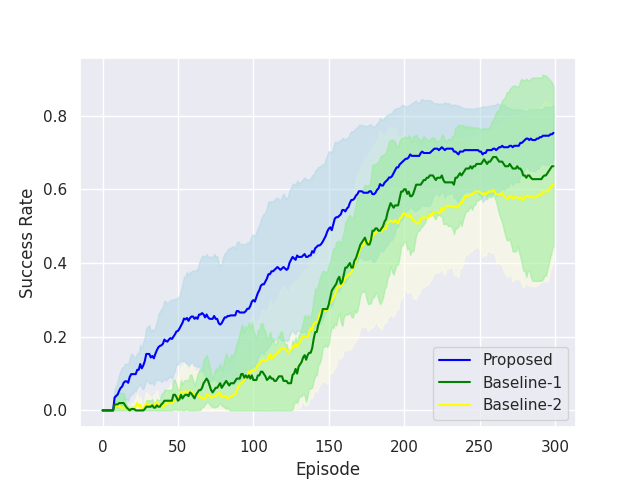}
    \label{fig:result_block_2}
    }
    \subfigure[Generalization]{
    \centering
    \includegraphics[width=1.35in]{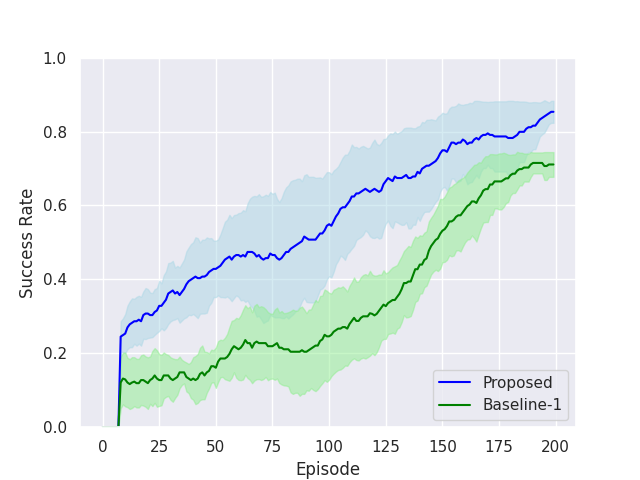}
    \label{fig:result_block_3}
    }

    \caption{The performance on the training and generalization, averaged on all the randomly generated tasks. }
    \label{fig:results}
    \vspace*{-10pt}
\end{figure}

\subsection{LTL Task Solving}
{\bf Training.} The performance comparisons of training, averaged over all the randomly generated tasks, are shown in Figure \ref{fig:results}. In all three domains, we can see clear advantage of the proposed framework over baselines. The Baseline-1 uses Q-learning to find the satisfying plan in the product MDP $\mathcal{P}$ and ground the operators by controller policies to finish the task. It is purely model-free and does not utilize the transition information of either the FSA $\mathcal{A}_{\phi}$ or the symbolic rules in $\Phi$. It directly uses trial-and-error to learn the precondition and effects of operators. Hence, its learning efficiency is low. The Baseline-2 uses Reward Machine to learn sub-policies for every automaton state in FSA, which decomposes the task into sub-tasks of reaching different automaton states. Based on the transitions in $\mathcal{A}_{\phi}$, it uses Q-learning to find a sequence of grounded operators (sub-policy) to reach every automaton state. However, this method separately solves every sub-task of transiting between automaton states, which may ignore the global optimality as shown in Figure \ref{fig:results}. In addition, some automaton states are not on the optimal paths from initial states to accepting states, and learning sub-policies to reach those states can reduce the learning efficiency. In contrast, the proposed framework uses a planner to find a plan of grounded operators to directly reach the accepting automaton state, which can keep the global optimality and ignore distracting states by value iterations. We can see that Baseline 2 performs worst, since it does not utilize the transition rules and suffers from sub-optimality resulted from task decomposition. 
\\
{\bf Generalization. } With the trained controller (low-level) policies and value functions, we compare the re-training steps of the proposed framework and Baseline-1 in unseen tasks. Since the task in Reward Machine is decomposed according to the FSA of the task formula $\phi$, we can not compare Baseline-2 with the proposed framework for generalization to unseen tasks. 
In Figure \ref{fig:results}, we can see that the proposed framework can generalize to new tasks significantly faster (more than 5x) than the Baseline-1, which shows the effect of learned symbolic transition rules and product MDP on improving the generalization capability. More results are presented in the Appendix of supplementary materials.

\section{Conclusion}
In this work, we propose a new learning framework for following temporal logic instructions in a relational domain that involves causal dependencies. Different from prior works, the agent does not know these causal dependencies as prior knowledge. We propose to use ILP-based method to learn symbolic operators and describe them as symbolic operators which can build a symbolic transition model for operators. Based this learned transition model, we can build a product MDP as the high-level abstraction of the environment, and the solve the given LTL task by a hierarchical RL approach. The advantage of the proposed framework is verified in three different domains.


\bibliographystyle{plain}
\bibliography{reference}

\begin{thebibliography}{10}

\bibitem{andreas2017modular}
Jacob Andreas, Dan Klein, and Sergey Levine.
\newblock Modular multitask reinforcement learning with policy sketches.
\newblock In {\em International Conference on Machine Learning}, pages
  166--175. PMLR, 2017.

\bibitem{araki2021logical}
Brandon Araki, Xiao Li, Kiran Vodrahalli, Jonathan DeCastro, Micah~J Fry, and
  Daniela Rus.
\newblock The logical options framework.
\newblock {\em arXiv preprint arXiv:2102.12571}, 2021.

\bibitem{araki2019learning}
Brandon Araki, Kiran Vodrahalli, Thomas Leech, Cristian-Ioan Vasile, Mark~D
  Donahue, and Daniela~L Rus.
\newblock Learning to plan with logical automata.
\newblock 2019.

\bibitem{arora2018review}
Ankuj Arora, Humbert Fiorino, Damien Pellier, Marc M{\'e}tivier, and Sylvie
  Pesty.
\newblock A review of learning planning action models.
\newblock {\em The Knowledge Engineering Review}, 33, 2018.

\bibitem{camacho2019ltl}
Alberto Camacho, Rodrigo~Toro Icarte, Toryn~Q Klassen, Richard~Anthony
  Valenzano, and Sheila~A McIlraith.
\newblock Ltl and beyond: Formal languages for reward function specification in
  reinforcement learning.
\newblock In {\em IJCAI}, volume~19, pages 6065--6073, 2019.

\bibitem{coumans2016pybullet}
Erwin Coumans and Yunfei Bai.
\newblock Pybullet, a python module for physics simulation for games, robotics
  and machine learning, 2016.
\newblock {\em URL http://pybullet. org}, 2016.

\bibitem{de2020temporal}
Giuseppe De~Giacomo, Marco Favorito, Luca Iocchi, Fabio Patrizi, and Alessandro
  Ronca.
\newblock Temporal logic monitoring rewards via transducers.
\newblock In {\em Proceedings of the International Conference on Principles of
  Knowledge Representation and Reasoning}, volume~17, pages 860--870, 2020.

\bibitem{de2020restraining}
Giuseppe De~Giacomo, Luca Iocchi, Marco Favorito, and Fabio Patrizi.
\newblock Restraining bolts for reinforcement learning agents.
\newblock In {\em Proceedings of the AAAI Conference on Artificial
  Intelligence}, volume~34, pages 13659--13662, 2020.

\bibitem{dong2019neural}
Honghua Dong, Jiayuan Mao, Tian Lin, Chong Wang, Lihong Li, and Denny Zhou.
\newblock Neural logic machines.
\newblock {\em arXiv preprint arXiv:1904.11694}, 2019.

\bibitem{duret2016spot}
Alexandre Duret-Lutz, Alexandre Lewkowicz, Amaury Fauchille, Thibaud Michaud,
  Etienne Renault, and Laurent Xu.
\newblock Spot 2.0—a framework for ltl and w-automata manipulation.
\newblock In {\em International Symposium on Automated Technology for
  Verification and Analysis}, pages 122--129. Springer, 2016.

\bibitem{eppe2019semantics}
Manfred Eppe, Phuong~DH Nguyen, and Stefan Wermter.
\newblock From semantics to execution: Integrating action planning with
  reinforcement learning for robotic causal problem-solving.
\newblock {\em Frontiers in Robotics and AI}, page 123, 2019.

\bibitem{evans2018learning}
Richard Evans and Edward Grefenstette.
\newblock Learning explanatory rules from noisy data.
\newblock {\em Journal of Artificial Intelligence Research}, 61:1--64, 2018.

\bibitem{florensa2018automatic}
Carlos Florensa, David Held, Xinyang Geng, and Pieter Abbeel.
\newblock Automatic goal generation for reinforcement learning agents.
\newblock In {\em International conference on machine learning}, pages
  1515--1528. PMLR, 2018.

\bibitem{fox2003pddl2}
Maria Fox and Derek Long.
\newblock Pddl2. 1: An extension to pddl for expressing temporal planning
  domains.
\newblock {\em Journal of artificial intelligence research}, 20:61--124, 2003.

\bibitem{hasanbeig2020deep}
Mohammadhosein Hasanbeig, Daniel Kroening, and Alessandro Abate.
\newblock Deep reinforcement learning with temporal logics.
\newblock In {\em International Conference on Formal Modeling and Analysis of
  Timed Systems}, pages 1--22. Springer, 2020.

\bibitem{he2015towards}
Keliang He, Morteza Lahijanian, Lydia~E Kavraki, and Moshe~Y Vardi.
\newblock Towards manipulation planning with temporal logic specifications.
\newblock In {\em 2015 IEEE international conference on robotics and automation
  (ICRA)}, pages 346--352. IEEE, 2015.

\bibitem{icarte2018using}
Rodrigo~Toro Icarte, Toryn Klassen, Richard Valenzano, and Sheila McIlraith.
\newblock Using reward machines for high-level task specification and
  decomposition in reinforcement learning.
\newblock In {\em International Conference on Machine Learning}, pages
  2107--2116. PMLR, 2018.

\bibitem{icarte2022reward}
Rodrigo~Toro Icarte, Toryn~Q Klassen, Richard Valenzano, and Sheila~A
  McIlraith.
\newblock Reward machines: Exploiting reward function structure in
  reinforcement learning.
\newblock {\em Journal of Artificial Intelligence Research}, 73:173--208, 2022.

\bibitem{illanes2020symbolic}
Le{\'o}n Illanes, Xi~Yan, Rodrigo~Toro Icarte, and Sheila~A McIlraith.
\newblock Symbolic plans as high-level instructions for reinforcement learning.
\newblock In {\em Proceedings of the International Conference on Automated
  Planning and Scheduling}, volume~30, pages 540--550, 2020.

\bibitem{jiang2020temporal}
Yuqian Jiang, Sudarshanan Bharadwaj, Bo~Wu, Rishi Shah, Ufuk Topcu, and Peter
  Stone.
\newblock Temporal-logic-based reward shaping for continuing learning tasks.
\newblock {\em arXiv preprint arXiv:2007.01498}, 2020.

\bibitem{jothimurugan2019composable}
Kishor Jothimurugan, Rajeev Alur, and Osbert Bastani.
\newblock A composable specification language for reinforcement learning tasks.
\newblock {\em Advances in Neural Information Processing Systems}, 32, 2019.

\bibitem{kokel2021reprel}
Harsha Kokel, Arjun Manoharan, Sriraam Natarajan, Balaraman Ravindran, and
  Prasad Tadepalli.
\newblock Reprel: Integrating relational planning and reinforcement learning
  for effective abstraction.
\newblock In {\em Proceedings of the International Conference on Automated
  Planning and Scheduling}, volume~31, pages 533--541, 2021.

\bibitem{koller2007introduction}
Daphne Koller, Nir Friedman, Sa{\v{s}}o D{\v{z}}eroski, Charles Sutton, Andrew
  McCallum, Avi Pfeffer, Pieter Abbeel, Ming-Fai Wong, David Heckerman, Chris
  Meek, et~al.
\newblock {\em Introduction to statistical relational learning}.
\newblock MIT press, 2007.

\bibitem{kress2018synthesis}
Hadas Kress-Gazit, Morteza Lahijanian, and Vasumathi Raman.
\newblock Synthesis for robots: Guarantees and feedback for robot behavior.
\newblock {\em Annual Review of Control, Robotics, and Autonomous Systems},
  1:211--236, 2018.

\bibitem{kupferman2001model}
Orna Kupferman and Moshe~Y Vardi.
\newblock Model checking of safety properties.
\newblock {\em Formal Methods in System Design}, 19(3):291--314, 2001.

\bibitem{leon2020systematic}
Borja~G Le{\'o}n, Murray Shanahan, and Francesco Belardinelli.
\newblock Systematic generalisation through task temporal logic and deep
  reinforcement learning.
\newblock {\em arXiv preprint arXiv:2006.08767}, 2020.

\bibitem{li2021reactive}
Shen Li, Daehyung Park, Yoonchang Sung, Julie~A Shah, and Nicholas Roy.
\newblock Reactive task and motion planning under temporal logic
  specifications.
\newblock {\em arXiv preprint arXiv:2103.14464}, 2021.

\bibitem{li2018policy}
Xiao Li, Yao Ma, and Calin Belta.
\newblock A policy search method for temporal logic specified reinforcement
  learning tasks.
\newblock In {\em 2018 Annual American Control Conference (ACC)}, pages
  240--245. IEEE, 2018.

\bibitem{littman2017environment}
Michael~L Littman, Ufuk Topcu, Jie Fu, Charles Isbell, Min Wen, and James
  MacGlashan.
\newblock Environment-independent task specifications via gltl.
\newblock {\em arXiv preprint arXiv:1704.04341}, 2017.

\bibitem{luketina2019survey}
Jelena Luketina, Nantas Nardelli, Gregory Farquhar, Jakob Foerster, Jacob
  Andreas, Edward Grefenstette, Shimon Whiteson, and Tim Rockt{\"a}schel.
\newblock A survey of reinforcement learning informed by natural language.
\newblock {\em arXiv preprint arXiv:1906.03926}, 2019.

\bibitem{maly2013iterative}
Matthew~R Maly, Morteza Lahijanian, Lydia~E Kavraki, Hadas Kress-Gazit, and
  Moshe~Y Vardi.
\newblock Iterative temporal motion planning for hybrid systems in partially
  unknown environments.
\newblock In {\em Proceedings of the 16th international conference on Hybrid
  systems: computation and control}, pages 353--362, 2013.

\bibitem{mccarthy1960programs}
John McCarthy et~al.
\newblock {\em Programs with common sense}.
\newblock RLE and MIT computation center Cambridge, MA, USA, 1960.

\bibitem{mnih2015human}
Volodymyr Mnih, Koray Kavukcuoglu, David Silver, Andrei~A Rusu, Joel Veness,
  Marc~G Bellemare, Alex Graves, Martin Riedmiller, Andreas~K Fidjeland, Georg
  Ostrovski, et~al.
\newblock Human-level control through deep reinforcement learning.
\newblock {\em nature}, 518(7540):529--533, 2015.

\bibitem{nair2018visual}
Ashvin~V Nair, Vitchyr Pong, Murtaza Dalal, Shikhar Bahl, Steven Lin, and
  Sergey Levine.
\newblock Visual reinforcement learning with imagined goals.
\newblock {\em Advances in Neural Information Processing Systems},
  31:9191--9200, 2018.

\bibitem{pasula2007learning}
Hanna~M Pasula, Luke~S Zettlemoyer, and Leslie~Pack Kaelbling.
\newblock Learning symbolic models of stochastic domains.
\newblock {\em Journal of Artificial Intelligence Research}, 29:309--352, 2007.

\bibitem{payani2019learning}
Ali Payani and Faramarz Fekri.
\newblock Learning algorithms via neural logic networks.
\newblock {\em arXiv preprint arXiv:1904.01554}, 2019.

\bibitem{rocktaschel2017end}
Tim Rockt{\"a}schel and Sebastian Riedel.
\newblock End-to-end differentiable proving.
\newblock In {\em Advances in Neural Information Processing Systems}, pages
  3788--3800, 2017.

\bibitem{silver2021learning}
Tom Silver, Rohan Chitnis, Joshua Tenenbaum, Leslie~Pack Kaelbling, and
  Tom{\'a}s Lozano-P{\'e}rez.
\newblock Learning symbolic operators for task and motion planning.
\newblock In {\em 2021 IEEE/RSJ International Conference on Intelligent Robots
  and Systems (IROS)}, pages 3182--3189. IEEE, 2021.

\bibitem{vaezipoor2021ltl2action}
Pashootan Vaezipoor, Andrew~C Li, Rodrigo A~Toro Icarte, and Sheila~A
  Mcilraith.
\newblock Ltl2action: Generalizing ltl instructions for multi-task rl.
\newblock In {\em International Conference on Machine Learning}, pages
  10497--10508. PMLR, 2021.

\bibitem{van2016deep}
Hado Van~Hasselt, Arthur Guez, and David Silver.
\newblock Deep reinforcement learning with double q-learning.
\newblock In {\em Proceedings of the AAAI conference on artificial
  intelligence}, volume~30, 2016.

\bibitem{van2015learning}
Herke Van~Hoof, Tucker Hermans, Gerhard Neumann, and Jan Peters.
\newblock Learning robot in-hand manipulation with tactile features.
\newblock In {\em 2015 IEEE-RAS 15th International Conference on Humanoid
  Robots (Humanoids)}, pages 121--127. IEEE, 2015.

\bibitem{vasile2013sampling}
Cristian~Ioan Vasile and Calin Belta.
\newblock Sampling-based temporal logic path planning.
\newblock In {\em 2013 IEEE/RSJ International Conference on Intelligent Robots
  and Systems}, pages 4817--4822. IEEE, 2013.

\bibitem{wang2021scalable}
Zhuo Wang, Wei Zhang, Ning Liu, and Jianyong Wang.
\newblock Scalable rule-based representation learning for interpretable
  classification.
\newblock {\em Advances in Neural Information Processing Systems}, 34, 2021.

\bibitem{wongpiromsarn2010receding}
Tichakorn Wongpiromsarn, Ufuk Topcu, and Richard~M Murray.
\newblock Receding horizon control for temporal logic specifications.
\newblock In {\em Proceedings of the 13th ACM international conference on
  Hybrid systems: computation and control}, pages 101--110, 2010.

\bibitem{xu2019transfer}
Zhe Xu and Ufuk Topcu.
\newblock Transfer of temporal logic formulas in reinforcement learning.
\newblock In {\em IJCAI: proceedings of the conference}, volume~28, page 4010.
  NIH Public Access, 2019.

\bibitem{yuan2019modular}
Lim~Zun Yuan, Mohammadhosein Hasanbeig, Alessandro Abate, and Daniel Kroening.
\newblock Modular deep reinforcement learning with temporal logic
  specifications.
\newblock {\em arXiv preprint arXiv:1909.11591}, 2019.

\end{thebibliography}

\renewcommand\thesection{\Alph{section}}

\onecolumn
\newpage

\section{Appendix}

\subsection{Task Specification: Co-safe Linear Temporal Logics}
\label{sec:ltl}
Co-safe LTL is a fragment of LTL which combines Boolean operators with temporal reasoning. Let $\mathcal{Q}$ be a set of Boolean atoms defined with symbolic MDP.
The syntax of co-safe LTL formula $\phi$ over $\mathcal{Q}$ can be inductively defined as below:
\begin{equation}
    \phi:=p | \neg\phi | \phi_1\vee\phi_2 | \bigcirc\phi | \phi_1\cup\phi_2 \nonumber
\end{equation}
where $p$ is a grounded predicate denoting the truth statement corresponding to a certain property of objects or events in the environment. Different from previous works on LTL in RL, we define the LTL over predicates rather than propositions, so that the relations of objects can be considered. Here $\neg$ is negation, $\vee$ is disjunction, $\bigcirc$ is "next", and $\cup$ is "until". The derived operators are conjunction ($\wedge$), "eventually" ($\Diamond\equiv\text{True}\cup\phi$) and "always" ($\Box\phi\equiv\neg\Diamond\neg\phi$). Specifically, $\phi_1\cup\phi_2$ means that $\phi_1$ is true until $\phi_2$ is true, $\Diamond\phi$ means that there is a time when $\phi$ becomes true, $\Box\phi$ means that $\phi$ is always true.

In general, the semantics of LTL formulas are defined over infinite words, which are infinite sequences of letters from the alphabet $2^{\mathcal{Q}}$. Nevertheless, each word that is accepted by a co-safe LTL formula can be detected by one of its finite prefixes \cite{kupferman2001model}. Thus, co-safe LTL is a desirable specification language to express robotic tasks that must be accomplished in finite time.

{\bf Finite State Automaton.} In order to check the satisfiability of $\phi$, we can construct a finite state automaton (FSA) that accepts precisely all the satisfying words of $\phi$ \cite{kupferman2001model}. This FSA is defined as $\mathcal{A}_{\phi}:=\langle \mathcal{Z}_a, z_{a,0}, \Sigma, T_a, \mathcal{F}_a \rangle$, where $\mathcal{Z}_a$ is the finite set of automaton states, $z_{a,0}$ is the initial state, and $\Sigma$ is the alphabet of the LTL formula (we have $\Sigma=2^{\mathcal{Q}}$). The letters of $\Sigma$ are the possible truth assignments of the atoms in $\mathcal{Q}$. Further, $T_a:\mathcal{Z}_a\times\Sigma\times\mathcal{Z}_a\to[0,1]$ is the automaton transition function. $\mathcal{F}_a$ is the accepting states of $\mathcal{A}_{\phi}$ ($\mathcal{F}_a\subset\mathcal{Z}_a$). The accepting runs of $\mathcal{A}_{\phi}$ are paths from $z_{a,0}$ to a state in $\mathcal{F}_a$ following the transition function $T_a$. The letters along the path represent the sequence of truth assignments of the atoms to satisfy the specification. Given the LTL formula $\phi$, the corresponding FSA can be easily obtained by applying heuristics such as SPOT \cite{duret2016spot}.

\subsection{Logic Neural Network}
We adopt logic neural network to learn the preconditions of symbolic operators \cite{wang2021scalable}. The first layer is the logic layer consisting of conjunctions of input predicates. The second layer is the linear layer for learning weights of possible rules. The number of nodes of the logic layer is chosen among $\{64, 128, 256, 512\}$. We use the cross-entropy loss during the training. The L2 regularization is adopted to control the model complexity, and the coefficient of the regularization term in the loss function is selected among $\{1e-3, 1e-4, \ldots, 1e-8, 0\}$. We use the Adam optimizer with the mini-batch size of $32$. The learning rate is selected among $\{5e-3, 3e-3, 1e-3, 5e-4, 3e-4, 1e-4\}$. The logic neural network is trained for 100 epochs.

\subsection{Algorithms}
\label{sec:app_alg}
\begin{algorithm}[ht]
\caption{Learning Low-level Controller Policies by SAC-HER}
\label{alg:sac-her}
\begin{algorithmic}[1]
\REQUIRE Environment MDP $\mathcal{M}$, actors and critics for controller policies in $\Pi=\{\pi^1,\ldots,\pi^K\}$;
\FOR{$\pi^k\in\Pi$}
\STATE Initialize policy $\pi^k$ and critics $Q^k$, and empty the replay buffer $\mathcal{B}^k$;
\FOR{$i=1,\ldots$}
\STATE Initialize the environment with a random goal;
\STATE Collect a trajectory by policy $\pi^k$, and store it into $\mathcal{B}^k$;
\STATE Sample a minibatch of transitions from $\mathcal{B}^k$, and relabel the goal of each transition by future strategy;
\STATE Update policy $\pi^k$ and critics $Q^k$ by SAC objectives;
\ENDFOR
\ENDFOR
\STATE {\bf Return} actor and critics for controllers in $\Pi$
\end{algorithmic}
\end{algorithm}

\begin{algorithm}[ht]
\caption{Learning Symbolic Transition Rules}
\label{alg:ilp}
\begin{algorithmic}[1]
\REQUIRE symbolic MDP $\tilde{\mathcal{M}}$, atom set $\mathcal{Q}$, subgoal atoms $\mathcal{Q}_G$ and operator predicates $\mathcal{OP}$;
\STATE Collect symbolic transition dataset $D$;
\STATE The dataset $D$ is partitioned according to operator predicate and lifted effects, i.e., $D=\bigcup_{\text{op}\in\mathcal{OP}, m=0,1,\ldots}D_{\text{op},m}$;
\STATE For every $D_{\text{op}, m}$, with $D_{\text{op}, 0}$ as negative examples, we train the logic neural network in \eqref{logic_layer} to learn the pre-condition rule;
\STATE For every operator $\text{op}\in\mathcal{OP}$, we compute the empirical distribution of lifted effects which correspond to the same pre-condition rule;
\STATE The symbolic transition model $\Phi$ can be composed by the pre-condition rules of operators and distributions of lifted effects;
\STATE {\bf Return} Symbolic transition model $\Phi$
\end{algorithmic}
\end{algorithm}

\subsection{Baseline Implementation}
\label{sec:base}
\subsubsection{Baseline 1: Q Learning}
In this method, we use Q-learning to learn the high-level policy for selecting operators. In every symbolic states, it uses $\epsilon$-greedy strategy to explore different operators and learn the causal dependencies by negative rewards received from unsuccessful trials of operators. It does not utilize any transition model. The benefit of this method is the simplicity of its implementation, but the disadvantage is the weak capability of generalization, since the transition rules are not utilized explicitly. The details of this baseline is shown in Algorithm \ref{alg:ql}. 

\begin{algorithm}[ht]
\caption{Baseline 1: Q-learning}
\label{alg:ql}
\begin{algorithmic}[1]
\REQUIRE Environment MDP $\mathcal{M}=\langle\mathcal{S}, \bm{s}_0, \mathcal{A}, \mathcal{T}\rangle$; \\
symbolic MDP $\tilde{\mathcal{M}}=\langle\tilde{\mathcal{S}}, \tilde{\mathcal{A}}, \tilde{s}_0, \tilde{T},\tilde{R},\gamma\rangle$; \\
The set of grounded predicates (atoms) $\mathcal{Q}$, the set of subgoal atoms $\mathcal{Q}_G\subset\mathcal{Q}$, the set of objects $\mathcal{Q}$, the set of operator predicates $\mathcal{OP}$, the labeling function $L:\mathcal{S}\to\tilde{\mathcal{S}}$, and the replay buffer $\mathcal{B}$; \\
The LTL formula $\phi$ for task specification; \\
The FSA of the given LTL formula $\mathcal{A}_{\phi}=\langle\mathcal{Z}_a, z_{a,0}, \Sigma, T_a, \mathcal{F}_a\rangle$; \\
Number of training episodes $m$, length of each episode $l$;\\
\STATE Apply Algorithm \ref{alg:sac-her} to learn controller policies in $\Pi$;
\STATE {\it //Find a high-level plan $h$ over the product MDP $\mathcal{P}$ by Q learning:}
\STATE Initialize $Q:\mathcal{Z}_a\times\tilde{\mathcal{S}}\times\tilde{\mathcal{A}}\to\mathbb{R}, V:\mathcal{Z}_a\times\tilde{\mathcal{S}}\to\mathbb{R}$ to $0$;
\FOR{$i=1,\ldots,m$}
\STATE Init FSA state $z_a\leftarrow0$, $s$ a random init state of $\mathcal{M}$ and the init symbolic state $\tilde{s}:=L(s)$
\FOR{$j=1,\ldots,l$}
\STATE Use $\epsilon$-greedy to select the symbolic action (grounded operator) $\tilde{a}$ over $Q(z_a,\tilde{s},\cdot)$
\STATE From $\Pi$, select the low-level policy corresponding to $\tilde{a}$ and use it to finish $\tilde{a}$
\STATE Obtain the next symbolic state $\tilde{s}'$ and environment reward $r$
\STATE Store the transition tuple $(\tilde{s}, \tilde{a}, \tilde{s}', r)$ into $\mathcal{B}$
\STATE Sample a minibatch from $\mathcal{B}$ and update Q and value functions $Q(z_a,\tilde{s},\tilde{a}), V(z_a, \tilde{s})$ by TD loss \cite{van2015learning,van2016deep}
\ENDFOR
\ENDFOR
\STATE Select the optimal high-level plan $h$ of $\tilde{a}$ over $Q(z_a,\tilde{s},\tilde{a})$
\STATE {\bf Return} $h, Q(z_a,\tilde{s},\tilde{a}), V(z_a, \tilde{s})$
\end{algorithmic}
\end{algorithm}

\subsubsection{Baseline 2: Reward Machine}
Since the task is specified by an LTL formula, the completion of a task can be formulated by a non-Markov reward function. So, in this baseline, we transform the LTL formula into the finite state automaton (FSA) and formulate it as a reward machine (RM) to solve the task, which is a popular method for LTL tasks recently proposed in \cite{icarte2018using,icarte2022reward}. Specifically, we choose the Q-Learning for Reward Machines (QRM) \cite{icarte2018using} as a way to exploit reward machine structure. The key idea of QRM is to reuse transition experience to simultaneously learn optimal behaviours for the different RM states. The QRM learns a separate Q-value function $q_z$ for every RM state $z\in\mathcal{Z}$. Formally, for any transition experience $(s,a,s')$, we can write as below
\begin{equation}
    q_z(s,a)\leftarrow\delta_r(z)(s,a,s') + \gamma\max_{a'\in\mathcal{A}}q_{\delta_z(z, L(s,a,s'))}(s',a') \nonumber
\end{equation}
where $\delta_z, \delta_r$ are RM state transition and reward functions, and $L$ is labeling function for a transition tuple. Since the QRM formulates separate Q-value functions for different RM states, based on the FSA of any LTL task, the original task can be decomposed into subtasks of reaching different RM states, and every subtask can be solved by Q learning separately. The causal dependencies can also be learned and followed when solving these subtasks. 

\begin{figure}[ht]
    \vspace*{-10pt}
    \centering
    \subfigure[Map]{
        \centering
        \includegraphics[width=1in]{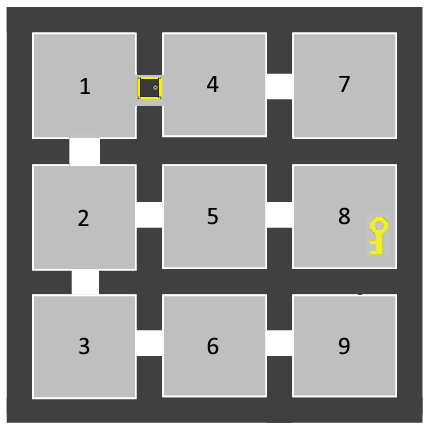}
        \label{fig:motivate_map}
    }
    \subfigure[Finite State Automaton]{
        \centering
        \includegraphics[width=2in]{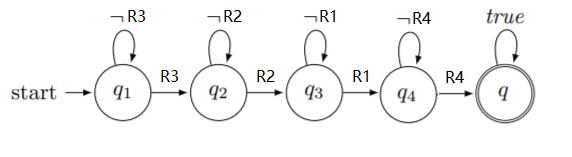}
        \label{fig:motivate_ltl}
    }
    \caption{Counter example. The atom R$k$ means visiting room $k$.}
    \label{fig:motivate}
    \vspace*{-10pt}
\end{figure}

However, the solutions learned by this QRM method can not be generalized from one task to another, since the Q-value function is defined for a specific RM state and the meaning of RM states is changed in another LTL task. Another deficiency of this QRM method is that, due to causal dependencies of operators, aggregating solutions locally optimal in subtasks cannot lead to a solution with global optimality. We have a example here. 
\\
{\bf Counter Example for QRM} With the map shown in Figure \ref{fig:motivate_map}, the agent is asked to visit several rooms in the order specified by an LTL formula, $\Diamond(R3\wedge\Diamond(R2\wedge\Diamond(R1\wedge\Diamond R4)))$, meaning that the agent should first visit room 3, then room 2, then room 1, and finally room 4. Note that there is a lock between room 1 and room 4, and the agent should first visit room 8 to pick up the key to open the lock. However, the user may not know this dependency in the beginning, and the given LTL does not contain the additional subtask of visiting room 8 to get the key. 
In QRM, the LTL task specification is first transformed into the automaton in Figure \ref{fig:motivate_ltl} which is used to define the reward machine, and the original task is decomposed into subtasks of room $3\to2, 2\to1$ and $1\to4$. The agent could learn to fetch the key in room 8 when solving the subtask of room $1\to4$. However, in the optimal solution with minimal number of steps, the agent should go to fetch the key in the subtask of room $2\to1$. This examples shows the sub-optimality of QRM. 

\subsubsection{Practical Implementation}
In reacher domain and block stacking domain, we implement the low-level policies for controllers as goal-conditioned policies which are trained with goal-conditioned critics, as presented in Section \ref{sec:op_to_act}. In block stacking domain, the low-level policies are not only pre-trained before executing any task, but also updated after trying every grounded operator (subtask) in the high-level plan. Since we separate the operation of picking up and placing the block, the low-level policies can reach a high successful probability ($>90\%$). However, in room domain, we simply use $32\times32$ MLP to realize the low-level polices for reaching four neighboring rooms, with the map of current room only as input, which are trained by deep Q-learning. In the high-level, before executing any task, the symbolic transition rules are learned by logic neural network in Section \ref{sec:symb_op}. 

In the baselines of Q-learning, we maintain a tabular Q function, $Q(z_a,\tilde{s},op)$, for every tuple of automaton state $z_a\in\mathcal{Z}_a$, symbolic state $\tilde{s}\in\tilde{\mathcal{S}}$ and grounded operator $\text{op}\in\tilde{\mathcal{A}}$, where $\epsilon$-greedy is used to select operator $op$ to try until any accepting state $z_{a,F}$ of automation $\mathcal{A}_{\phi}$ is reached. The advantage of this method is that it does not need to know the transition functions of relational MDP $\tilde{\mathcal{M}}$ and FSA $\mathcal{A}_{\phi}$, but it may waste a lot of samples to learn to satisfy the preconditions of operators and have weak generalization capability to other tasks with different FSA $\mathcal{A}_{\phi}$.

In the baseline of reward machine,  we adopt the Q-learning for Reward Machine (QRM) \cite{icarte2018using} which learns one subpolicy per state in $\mathcal{A}_{\phi}$ and uses off-policy learning to train each subpolicy in parallel. For example, if the task formula in reacher domain is $\phi=\Diamond\text{Visit}(r)\wedge\Diamond\text{Visit}(b)\wedge\Diamond\text{Visit}(y)$, the robot first finds a sequence of operators for reaching red ball $r$, then going to blue ball $b$ and finally reaching yellow ball $y$. 

\subsection{More Experiment Results}

\begin{figure}[H]
    \centering
    \subfigure[Room]{
    \centering
    \includegraphics[width=.75in, height=.7in]{maze_test1.png}
    \label{fig:result_room}
    }
    \subfigure[Sequence]{
    \centering
    \includegraphics[width=1.in]{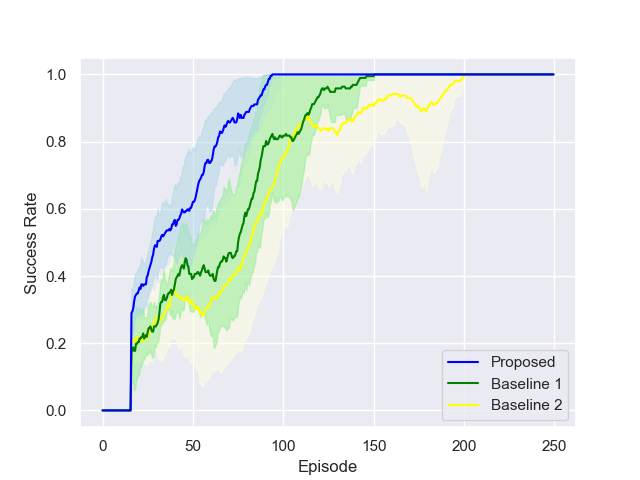}
    \label{fig:result_room_seq}
    }
    \subfigure[Or]{
    \centering
    \includegraphics[width=1.in]{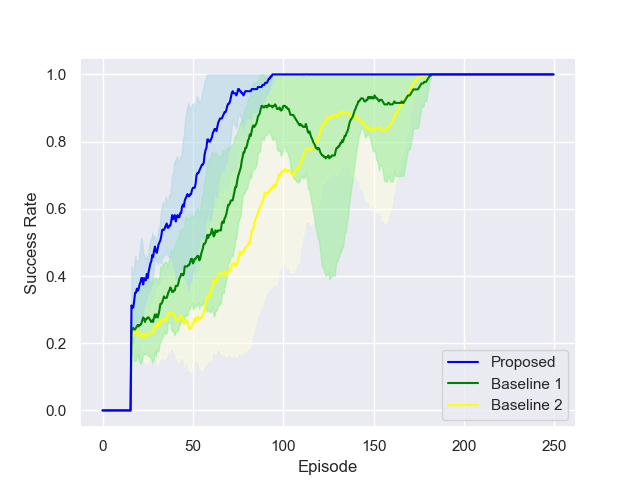}
    \label{fig:result_room_or}
    }
    \subfigure[Recursive]{
    \centering
    \includegraphics[width=1.in]{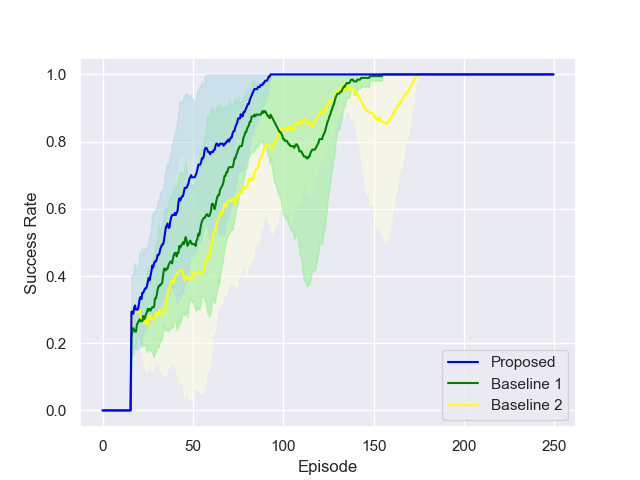}
    \label{fig:result_room_recur}
    }

    \subfigure[Reacher]{
    \centering
    \includegraphics[width=.75in, height=.7in]{reacher_img1.png}
    \label{fig:result_reacher}
    }
    \subfigure[Sequence]{
    \centering
    \includegraphics[width=1.in]{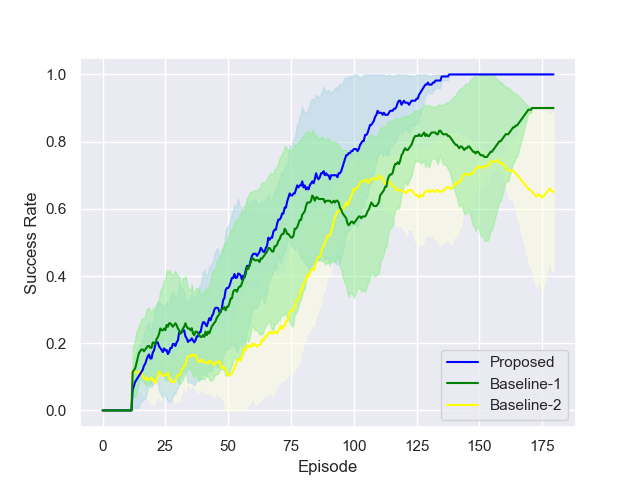}
    \label{fig:result_reacher_seq}
    }
    \subfigure[Or]{
    \centering
    \includegraphics[width=1.in]{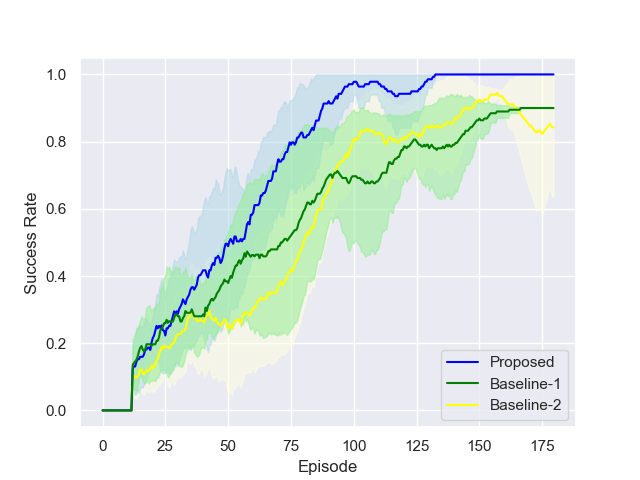}
    \label{fig:result_reacher_or}
    }
    \subfigure[Recursive]{
    \centering
    \includegraphics[width=1.in]{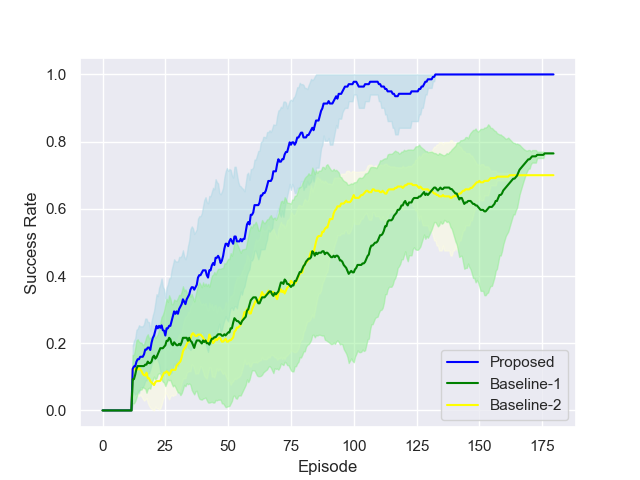}
    \label{fig:result_reacher_rec}
    }

    \subfigure[Block Stacking]{
    \centering
    \includegraphics[width=.75in, height=.7in]{block_img1.png}
    \label{fig:result_block}
    }
    \subfigure[Sequence]{
    \centering
    \includegraphics[width=1.in]{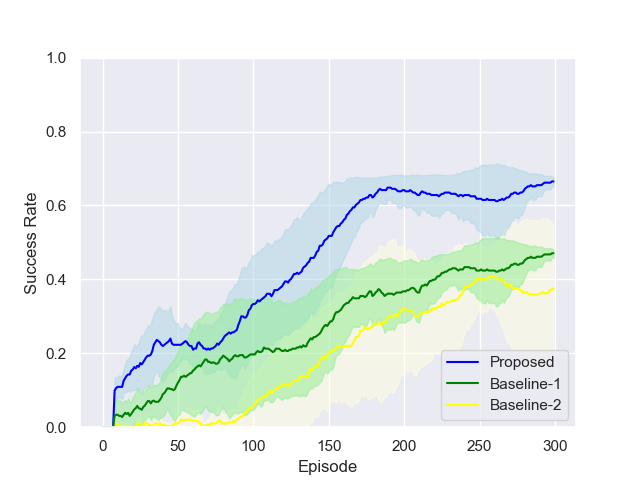}
    \label{fig:result_block_seq}
    }
    \subfigure[Or]{
    \centering
    \includegraphics[width=1.in]{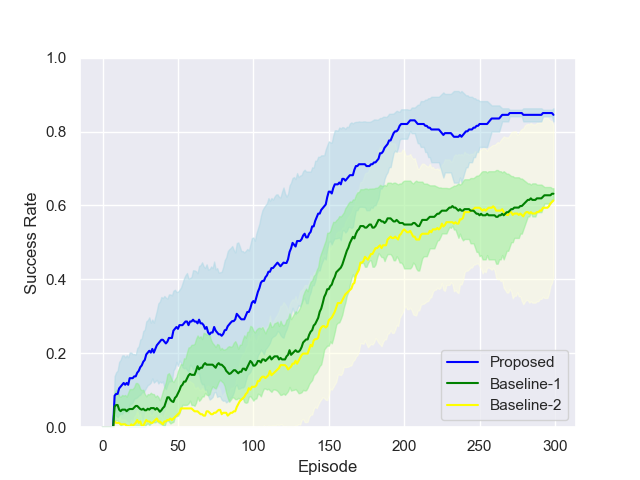}
    \label{fig:result_block_or}
    }
    \subfigure[Recursive]{
    \centering
    \includegraphics[width=1.in]{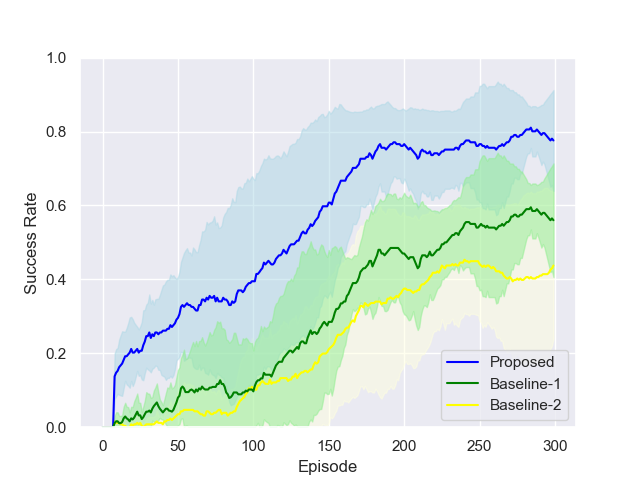}
    \label{fig:result_block_recur}
    }

    \caption{The performance on the training with results shown for all kinds of tasks. We can see that, the performance of Reward Machine (Baseline 2) is the worst, since it decomposes the original task and solves it separately. The proposed framework which uses learned transition rules performs the best in all cases. The sequence task is more difficult to learn than or task, since it needs more steps to finish than or tasks. The recursive task is the most difficult, since it has most complex logic structures.}
    \label{fig:results_more}
\end{figure}

\end{document}